\documentclass[letterpaper, 10 pt, conference]{ieeeconf}  % Comment this line out if you need a4paper

\IEEEoverridecommandlockouts                              % This command is only needed if 
\usepackage{graphics} % for pdf, bitmapped graphics files
\usepackage{epsfig} % for postscript graphics files
\usepackage{mathptmx} % assumes new font selection scheme installed
\usepackage{times} % assumes new font selection scheme installed
\usepackage{amsmath} % assumes amsmath package installed
\usepackage{amssymb}  % assumes amsmath package installed

\usepackage{algorithm}
\usepackage[noend]{algpseudocode}
\usepackage{setspace}
\usepackage{breqn}
\usepackage{color}
\usepackage{array}
\usepackage{tabu}
\usepackage{tabulary}
\newcolumntype{P}[1]{>{\centering\arraybackslash}m{#1}}
\usepackage{caption}
\usepackage{graphicx, subfigure}

\title{\LARGE \bf
Towards Planning and Control of \\ Hybrid Systems with Limit Cycle using LQR Trees}

\author{Ramkumar Natarajan$^{1,2}$ \hspace{2cm}Siddharthan Rajasekaran$^{1,2}$ \hspace{2cm} Jonathan D. Taylor$^3$% <-this % stops a space
\thanks{$^{1}$This work was equally contributed by these two authors. \hspace{3.5cm}$^{2}$ Ramkumar Natarajan {\tt\small <rnatarajan@wpi.edu>} and Siddharthan Rajasekaran {\tt\small <sperundurairajas@wpi.edu>}, Robotics Engineering,
Worcester Polytechnic Institute, Worcester, MA 01609. \hspace{7cm} $^{3}$Dr. Jonathan D. Taylor {\tt\small<jdtaylor@andrew.cmu.edu>}, Mechanical Engineering, Carnegie Mellon University, Pittsburgh, PA 15213}%
}

\begin{document}

\maketitle
\thispagestyle{empty}
\pagestyle{empty}

%%%%%%%%%%%%%%%%%%%%%%%%%%%%%%%%%%%%%%%%%%%%%%%%%%%%%%%%%%%%%%%%%%%%%%%%%%%%%%%%
\begin{abstract}

We present a multi-query recovery policy for a hybrid system with goal limit cycle. The sample trajectories and the hybrid limit cycle of the dynamical system are stabilized using locally valid Time Varying LQR controller policies which probabilistically cover a bounded region of state space. The original LQR Tree algorithm builds such trees for non-linear static and non-hybrid systems like a pendulum or a cart-pole. We leverage the idea of LQR trees to plan with a continuous control set, unlike methods that rely on discretization like dynamic programming to plan for hybrid dynamical systems where it is hard to capture the exact event of discrete transition. We test the algorithm on a compass gait model by stabilizing a dynamic walking hybrid limit cycle with point foot contact from random initial conditions. We show results from the simulation where the system comes back to a stable behavior with initial position or velocity perturbation and noise.

% We present a method that probabilistically covers a bounded region of state space with feasible sample trajectories that lead to a dynamically stable walking limit cycle behavior for hybrid systems. The sample trajectories and the dynamic walking limit cycle are stabilized using locally valid Time Varying LQR controller policies. The original LQR Tree algorithm builds such trees for highly non-linear static systems like pendulum and cart-pole. We leverage the idea of LQR trees to plan with a continuous control set unlike methods that rely on discretization like dynamic programming to plan for hybrid dynamical systems where it is hard to capture the exact event of discrete transition. We test the algorithm on a simple compass gait model by stabilizing a dynamic walking limit cycle with point foot contact from random initial conditions. We show results from simulation where the system comes back to a stable behavior with initial perturbation and noise. 

\end{abstract}

% Accessing man-made envi

%%%%%%%%%%%%%%%%%%%%%%%%%%%%%%%%%%%%%%%%%%%%%%%%%%%%%%%%%%%%%%%%%%%%%%%%%%%%%%%%
\section{INTRODUCTION}
% Robots are becoming more common in human environments. It is, however, difficult for current robots to access all man-made environments constructed for gait like locomotion. Solving this problem often involves controlling highly nonlinear and hybrid systems. Often controlling robots with periodic gaits involve 

% Building robots to access man-made environments and enable friendly interaction with humans has been a great challenge. Bipedal walking robots are generally considered the most suitable design to interact with the existing built world, but such systems are highly nonlinear with hybrid dynamics posing a challenge to conventional control designs. 
Bipedal walking robots can easily access man-made environments and enable friendly interaction with humans, but such systems are highly nonlinear with hybrid dynamics posing a challenge to conventional control designs. Controllers designed to track a trajectory for highly nonlinear systems provide only a reflex policy and can at most succeed only in a local region of attraction around the target trajectory being tracked. To find a policy from any given initial state one can discretize the state space and use dynamic programming. However in the problem of stabilizing a hybrid system, which has continuous dynamics punctuated by discrete transitions, one must discretize the system using a fine resolution to capture the jump event with reasonable tolerance. An approach to control a compass gait where many approximations are made to adapt the existing methods to hybrid systems is described in \cite{DP}.

Also, there is a large variety of robots that consist of a logical discrete event decision-making system interacting with a continuous time process. The action capabilities of such robots cannot be captured with a single controller. These type of hybrid systems have multiple system dynamics governing their behavior and can therefore have multiple discontinuities in any of their trajectory in the state space. With the increase in the development of highly sophisticated robots, the need for a hybrid controller is imperative. Our work focuses on extending a relatively new and a continuous method of controller design to such hybrid systems. 

This approach is inspired by randomized feedback motion planning methods, which cover a bounded region of the state space with locally valid linear quadratic regulator (LQR) policies \cite{lqrtree} that lead to a target trajectory. Tedrake \cite{lqrtree} has mentioned LQR trees as a scalable algorithm that could avoid the pitfall of discretization in dynamic programming and also use continuous control actions. However, the original algorithm did not consider a hybrid system in its formulation and was tested only on a relatively simpler platform like a pendulum. Also, the original method only stabilizes the region of attraction of LQR controller around a stabilizable fixed point. We focus on extending the LQR trees to stabilize a periodic limit cycle of a hybrid system. That is, we would cover a bounded region of state space with funnels around sample trajectories that lead to the region of attraction around the goal limit cycle of a hybrid system. The funnels are regions of attraction around a trajectory created by a controller that act as a vacuum tube and sucks any state inside the tube so that they do not leave the tube while driving the state towards the end of the trajectory. These regions of attraction can be found using Lyapunov function verification. Tedrake \cite{sosalg} shows that recent methods that use Sums-of-Squares (SoS) optimization to verify Lyapunov functions allow very fast determination of these regions of attraction. 

Reist \textit{et. al}, \cite{lqrtree2} provide a very similar algorithm which verifies the funnels empirically through simulation instead of using SoS optimization. This method mainly aims at simplifying the implementation of the verification process at the cost of time complexity. It is tested on a more complex, 2 DoF cart-pole system. Here the size of a funnel is initialized to a very high value and is trimmed for every sample that fails to reach the goal under the time varying LQR policy of the nominal trajectory. Hence the method requires a huge number of samples, both to build the tree and to verify funnels. However, the number of nodes in the tree and several other results will be unchanged even while using the formal verification. Hence we will be comparing our results to this implementation.\footnote{The original implementation only shows results tested on a pendulum. A 2-DoF cart-pole system would be better to compare with since a compass gait is a 2 DoF system too.}

The original LQR tree algorithm ensures that any sequence of such funnels ends at the region of attraction of the final controller that stabilizes the system at a fixed point. In this paper, we design sequences of funnels that end at the goal limit cycle. For this, we ensure that the size of each funnel, at their end, which directly connects to the limit cycle is less than the funnel of the limit cycle itself. Also, we use trajectory optimization to generate hybrid trajectories instead of a single continuous trajectory which enables our algorithm to work for hybrid systems. 

%Posa \textit{et. al,} \cite{posa} presents a direct trajectory optimization method for rigid bodies with contact in which they let the optimizer figure out the best sequence of modes the hybrid system should operate in. However, for point to point trajectory planning we assume that a trajectory exists with at most two mode sequences. With this assumption, we successfully implement LQR Trees for the hybrid compass gait and test the robustness with both velocity and position perturbations along with noise in the forward simulation of the system.

\section{RELATED WORK}

% LQR trees are very similar to RRT \cite{rrt} except that they 1) plan a trajectory in state space using direct collocation \cite{underactuated} instead of the extend operation in RRT and 2) use the controller cost function for finding the nearest node. In this way, the algorithm includes the controller constraints in the planning phase apart from yielding a feasible policy. 
 %We investigate the dynamic walking using compass gait which was first modeled by McGeer \cite{cgait}. The success of Tedrake's LQR Trees \cite{lqrtree} motivates our interest to extend it to a hybrid system like compass gait which has a goal trajectory\footnote{The original method was designed for a goal state}. The idea of LQR trees is to cover the stabilizable state space with locally feasible trajectories that serve as a lookup table. The limit cycle is stabilized using the Time Varying Linear Quadratic Regulator (TVLQR) controller. Manchester \textit{et. al} \cite{hybrid} provide a method to find the region of attraction around the limit cycle by decomposing the dynamics around it into tangential and transverse components. The authors then find a Lyapunov candidate in the transverse dynamics using SoS analysis \cite{sos}. 

Stabilizing the gait has been a great interest among robotics researchers and falls into two broad classes: 1) Based on the Zero Moment Pole (ZMP) principle 2) Passive dynamic and limit cycle walkers. Robots like Honda ASIMO uses the ZMP technique \cite{hondazmp} to keep the dynamical degrees of freedom fully actuated. The criteria in ZMP is to keep the center of pressure within the support polygon of the stance foot. However, the motions resulting from these techniques are very unnatural and inefficient. 

There is a large body of work that aims to find the most stable limit cycle. Dai \textit{et al.} \cite{robustlimit} formulates an optimization problem to find the limit cycle that maximizes its robustness against external disturbances. They do this by minimizing the mean controller costs of limit cycle states encountering collision so that the reset states post impact lie close to the limit cycle. But our method is indifferent to any such goal limit cycle and brings the system back to a given goal limit cycle trajectory from far outside the region of attraction. 
 
To deal with states that are far outside this region of attraction, \cite{rrtcontrol} and \cite{rgrrt} plan trajectories beforehand which can act as look-up table that provides a policy for any given initial state. The recent trend in verifying a Lyapunov function using convex optimization techniques gave birth to several effective methods like \cite{lqrtree} and \cite{lqrtree2} that did not exist before due to the time overhead in verifying Lyapunov functions. Numerical methods for computing regions of finite-time invariance \cite{funnels} (``verification of funnels") around solutions of polynomial differential equations is extensively used in this paper. In fact, the idea is to cover the state space with the funnels around the sample trajectories that lead to the region of attraction of the goal trajectory. These funnels have non-zero volumes in state space and hence can effectively cover a bounded region filled with nominal trajectories
. Moreover, it takes relatively less number of funnels to cover the entire region of state space compared to the number of nodes in other methods like Probabilistic Road Map (PRM) \cite{prm} to cover a given region.

There are several previous works that describe the usage of sample paths as fundamental representation of policies. Initial attempts to use sampling based planner to control nonlinear hybrid systems uses Rapidly-exploring Random Tree (RRT) \cite{rrtcontrol}. Here, the nearest state in the tree to a state sampled at random is forward simulated with a random control input. Thus, due to forward simulation, every edge in the tree is a feasible trajectory with which we have a policy to go from the start node to any other node in the tree (which may include goal if a path is feasible). Most of the work in this field following \cite{rrtcontrol} can be categorized as improving the 1) Sampling distribution \cite{rgrrt} 2) Distance metric \cite{lqrtree,lqrtree2,learningdist}  and 3) Extend operation \cite{lqrtree,lqrtree2}. LQR Tree algorithm \cite{lqrtree} uses the controller cost function as distance metric which improves the success rate of finding a trajectory using direct collocation \cite{dircol} from the nearest point on the tree to the sampled point (extend operation). Our work focuses on extending the capability of LQR Trees to stabilize a hybrid trajectory. 

The remainder of this paper is outlined as follows: Section III sets the mathematical premise for the proposed extension. Section IV explains the proposed method in detail, viz., the basic principle of Time Varying Linear Quadratic Regulator (TVLQR) and how it is used to stabilize a goal limit cycle, the estimation of the limit cycle itself, the details involved in using the direct collocation trajectory optimization for hybrid trajectories and putting it together to cover the bounded region of state space with recovery policies. The testbed and the collision dynamics that is responsible for the discrete jump in the state space are described in section V. The algorithm is experimentally evaluated using the simulation of compass gait in section VI. Section VII and VIII offer a discussion and concluding remarks.

\section{PROBLEM STATEMENT}

Consider a hybrid system with continuous dynamics $\dot{\textbf{x}} = \textbf{f}(\textbf{x},\textbf{u})$ and discrete transitions at $\textbf{x}_{guard}$ with stabilizable goal limit cycle given by $\{\textbf{x}_{0}^{l}(t), \textbf{u}_{0}^{l}(t)\}$ where $\textbf{x}_{0}^{l}(t)$ is the limit cycle trajectory and $\textbf{u}_{0}^{l}(t)$ is the open loop control law. Let $\mathbb{X}$ be the entire state space of this system and $\mathbb{X}_S$ be the set of stabilizable states of this space. The LQR tree algorithm has a number of feedback stabilized sample trajectories which we'll denote using $\{\textbf{x}^i(t), \textbf{u}^i(t)\}$ for $i^{th}$ trajectory in the tree. Let each trajectory start at time $t_{0}^i$ and end at time $t_{f}^i$. We use a controller $c^i$ to stabilize the system around every $i^{th}$ trajectory which results in a region of attraction $F^i$ around this trajectory such that $F^i \in \mathbb{X}_S$. From the LQR trees for any $i$ we have, $\textbf{x}^i(t_{f}^i) \in F^j$ for some $j$. We design the system such that, the same holds for every trajectory and the system always ends in the respective parent's funnel at the end of each child's trajectory ultimately leading to the funnel of goal limit cycle, $F^l$. Our primary objective is that, as the number of sample trajectories increase, the union of all the funnels cover the entire stabilizable state space by also accounting for the discrete state transitions \textit{i.e.,} $\Bigg(\lim_{n \to \infty}  \bigcup\limits_{i=1}^{n} F^{i} \Bigg) \setminus \mathbb{X}_S = \emptyset$.

% Let us denote the entire state space by $\mathbb{X}$ and the set of stabilizable space by $\mathbb{X}_S$. Let $\{x_0^l,u_0^l\}$ be the stabilized goal limit cycle trajectory. Denoting the $i^{th}$ trajectory in the LQRTree by $\{x_i(t),u_i(t)\}$ with  starting $t_{i0}$ and final times $t_{if}$ and let $F_i$ be the funnel around it and $X_i$ denote the set of points that fall in the funnel. Our primary objective is that, as the number of samples increase the union of all the funnels covers the entire state space. Mathematically,\begin{align}
% \lim_{n \to \infty}  \bigcup\limits_{i=1}^{n} X_{i} - X_S = \emptyset
% \end{align}
% Also, one trajectory ends in another trajectory and eventually back to the goal limit cycle trajectory. Mathematically, for any $i^{th}$ trajectory,there exists a time $t_{i+1}$ corresponding to $x_{i+1}$ trajectory such that $x_i(t_{fi}) = x_{i+1}(t_{i+1}))$. In a similar way, there exists a set of trajectories such that for each trajectory denoted by $g$, there exists a time t such that,  $x_{g}(t_{fg}) = x_0^l(t)$. The above expression shows that    

\section{METHOD}

\subsection{TVLQR feedback stabilization for hybrid systems} \label{sec:tvlqr}

We use a TVLQR to stabilize the system around a given trajectory. We first explain the theory behind TVLQR controller which is necessary to estimate the funnels. Let us consider the sub-problem of designing a time-varying LQR feedback based on a time-varying linearization along a nominal trajectory.  Consider a smoothly differentiable, nonlinear system $\dot{\textbf{x}} = \textbf{f}(\textbf{x}, \textbf{u})$ with stabilizable limit cycle trajectory, $\textbf{x}_0^l(t)$ and $\textbf{u}_0^l(t)$. Let $\textbf{A}(t) = \frac{\partial \textbf{f}}{\partial \textbf{x}}  \Big|_t$, $\textbf{B}(t) = \frac{\partial \textbf{f}}{\partial \textbf{u}}  \Big|_t$ be the linearization of system dynamics with respect to state and input respectively. For now, assume we have the optimal cost-to-go matrix of TVLQR controller (around the limit cycle) $\textbf{S}^l(t)$ for the limit cycle trajectory. Determination of $\textbf{S}^l(t)$ is discussed later in this subsection.

The optimal cost-to-go for any nominal trajectory $\{ \textbf{x}_0(t), \textbf{u}_0(t) \}$ of the controller is given by \cite{lqrtree} as,
\begin{align}
\label{opt_cost}
J^*(\bar{\textbf{x}}, t) = \bar{\textbf{x}}^T(t)\textbf{S}(t)\bar{\textbf{x}}(t), \ \ \textbf{S}(t) = \textbf{S}^T(t)
\end{align}where $\textbf{S}(t)$ is the solution to the Riccati equation\begin{equation}
-\dot{\textbf{S}} = \textbf{Q} - \textbf{S}\textbf{B}\textbf{R}^{-1}\textbf{B}^T\textbf{S}+\textbf{S}\textbf{A}+\textbf{A}^T\textbf{S}
\label{riccati}
\end{equation}
and the optimal feedback policy of TVLQR controller is
\begin{align}
\label{optimal_u}
\bar{\textbf{u}}^*(t) = -\textbf{R}^{-1}\textbf{B}^T(t)\textbf{S}(t)\bar{\textbf{x}}(t) = -\textbf{K}(t)\bar{\textbf{x}}(t)
\end{align}
where $\bar{\textbf{x}}(t)$, $\bar{\textbf{u}}(t)$ are the state and input deviations from the nominal values $\textbf{x}_0(t)$ and $\textbf{u}_0(t)$,  $\textbf{Q}$ and $\textbf{R}$ penalize $\bar{\textbf{x}}(t)$ and $\bar{\textbf{u}}(t)$ respectively in the LQR cost function. From the above description of obtaining $\textbf{S}(t)$ for a trajectory we can conclude - 1) We need the cost to go matrix at final time step, $\textbf{Q}_f = \textbf{S}(t_f)$ from where we can integrate backwards to obtain $\textbf{S}(t)$. 2) The controller thus obtained promises only to put the system finally within an ellipsoid $\bar{\textbf{x}}^T(t_f)\textbf{S}(t_f)\bar{\textbf{x}}(t_f) < \rho(t_f)$ (intuitively this means the cost-to-go is less than some threshold $\rho(t_f)$) for some $\rho(t_f)$. The nominal trajectory $\{ \textbf{x}_{0}^i(t)$, $\textbf{u}_{0}^i(t) \}$ stabilized by a controller $c^{i}$ terminates at a new nominal trajectory  $\{ \textbf{x}^{i+1}_0(t)$, $\textbf{u}^{i+1}_0(t) \}$ stabilized by a controller $c^{i+1}$. It is required by condition 2) that $c^{i+1}$ must be able to stabilize any state in the ellipsoid resulted by applying $c^i$. The same holds for trajectories following $\{ \textbf{x}^{i+1}_0(t)$, $\textbf{u}^{i+1}_0(t) \}$. As we keep transitioning the system from one trajectory to another, the system eventually must find itself in a stabilizable state or the limit cycle. 

Consider a limit cycle that starts at $t_0$ and ends at $t_f$. Since the limit cycle ends at its start \textit{i.e.,} $\textbf{x}_0^l(t_0) = \textbf{x}_0^l(t_f)$, we also have $\textbf{S}^l(t_0) = \textbf{S}^l(t_f)$. One of the methods to find $\textbf{S}^l(t)$ is to initialize $\textbf{S}^l(t_f)$ to some arbitrary $\textbf{S}_0(t_f)$  \footnote{In $\textbf{S}_k(t_f)$, $k$ is the number of iterations. We can initialize it to penalize the state dimensions we care about. A good choice would be $\textbf{Q}$ from TVLQR cost function} and integrate backwards to get $\textbf{S}_0(t_0)$. In the next iteration, we re-initialize $\textbf{S}_1(t_f)$ to $\textbf{S}_0(t_0)$ and integrate backwards to get $\textbf{S}_1(t_0)$. This process is repeated till convergence \textit{i.e.,} till $\textbf{S}_k(t_0) = \textbf{S}_k(t_f)$ \footnote{In practice we would want them to be close under some tolerance.} for some $k$. The value of time varying cost to go matrix $\textbf{S}^l(t)$ is initialized to this $\textbf{S}_k(t)$. This is explained in Alg. \ref{lim}.

For a hybrid system, such as the compass gait, we have continuous dynamics and discrete transitions. Hence we cannot obtain $\textbf{S}(t)$ from a single Riccati equation since a solution to a first order quadratic differential equation must be smoothly differentiable. Therefore we have different Riccati equations for different modes the system is operating in. And we have a discrete `jump' event in the Riccati equation where we jump from one equation to another. This is called the jump Riccati equation \cite{hybrid}. The collision dynamics of the system is a function $\textbf{CD}^{pq}(\textbf{x})$ that maps states in mode $p$ just before the collision event (guards that cause discrete transitions) to states in mode $q$ just after the collision. That is, $\textbf{x}(t^+) = \textbf{CD}^{pq}(\textbf{x}(t^-))$, where $t^-$ and $t^+$ are the instances just before and after the collision event. We linearize the collision dynamics of the system to get $\textbf{A}_{cd} = \frac{\partial{\textbf{CD}}^{pq}}{\partial{x}}$. The cost to go matrix during the jump event is given by \cite{hybrid}, 
\begin{align}
\textbf{S}(t^-) = \textbf{A}^T_{cd}\textbf{S}(t^+)\textbf{A}_{cd}
\label{jump}
\end{align}

\subsection{Funnel around a trajectory}
Consider a system $\dot{\textbf{x}}=f(\textbf{x},\textbf{u})$ with a closed loop limit cycle $\textbf{x}_0^l(t), \textbf{u}_0^l(t)$ whose region of attraction is given by $\bar{\textbf{x}}^T(t)\textbf{S}^l(t)\bar{\textbf{x}}(t) < \rho^l(t)$, where $\bar{\textbf{x}}(t) =\textbf{x}(t) - \textbf{x}_0^l(t)$. $\textbf{S}^l(t)$ can be determined empirically as described in subsection \ref{sec:tvlqr}. Let $\zeta$ be a trajectory that takes the system from an arbitrary start state to the state $\textbf{x}_0^l(t_e)$ of the limit cycle. The closed loop limit cycle requires $\zeta$ to end at a state $\textbf{x}_e$ such that $(\textbf{x}_e - \textbf{x}_0^l(t_e))^T\textbf{S}^l(t_e)(\textbf{x}_e - \textbf{x}_0^l(t_e)) < \rho^l(t_e)$. We can view this as the allowed uncertainty at the tail end of the trajectory $\zeta$. Hence for a time varying system, instead of defining a discrete region of attraction at every time step, we define funnel \footnote{We use the term 'funnel' for a trajectory which is analogous to basin of attraction for a stabilizable state.} as the region around the trajectory where any point is guaranteed to be led by the closed loop system to the region of allowed uncertainty at the end of the trajectory. By stabilizing the system around a open loop trajectory $\textbf{x}_0(t), \textbf{u}_0(t)$, the TVLQR control design would give us the time varying controller $\textbf{u}^*(t)=\textbf{u}_0(t) - \textbf{K}(t)(\textbf{x}(t) - \textbf{x}_0(t))$ and also the cost-to-go function 
$J^*({\bar{\textbf{x}}},t)={\bar{\textbf{x}}}^{T}(t)\textbf{S}(t){\bar{\textbf{x}}}(t)$. This cost-to-go function is a candidate Lyapunov function for our system locally.

% Consider a system $\dot{\textbf{x}}=f(\textbf{x},\textbf{u})$ with equilibrium point at $\textbf{x}_g,\textbf{u}_g$. When we define a controller for this system, the region around the equilibrium point where any point can be stabilized to the equilibrium point is called the region of attraction for that system with the corresponding controller.
% %We estimate this basin of attraction by verifying that controller stabilizes the system to the equilibrium point by demonstrating that the function $V(x)$ is  a valid Lyapunov function over a bounded region of state space $\beta$, defined by
% %$$ \beta (\rho) : \{x | 0 \le V(x) \le \rho \}  $$
% %where \rho is a positive scalar. The
% However for a time varying system with $\dot{\textbf{x}}=A(t)\textbf{x}+B(t)\textbf{u}$, it is not possible to define a discrete region of attraction at every time step around the trajectory. For such a time varying system, funnel\footnote{We use the term 'funnel' for a trajectory which is analogous to basin of attraction for a stabilizable state} is defined as the region around the trajectory where any point is guaranteed to ultimately lead the closed loop system to the region of attraction of the goal. Once the system reaches the region of attraction of the goal the time invariant controller can stabilize the system to the equilibrium point.

Mathematically, we can define the funnel as the time varying region $\mathcal{B}$ ,where
\begin{align}
\mathcal{B}(t) = \{ \textbf{x} | F(\textbf{x},t) \in \beta^l\}  
\label{fun}
\end{align}where $F(\textbf{x},t)$ is the function that forward simulates the closed loop trajectory from $t$ to $t_f$ and $\beta^l$ is the region of attraction around the goal trajectory.

For any such trajectory $\zeta$, we use the cost-to-go, which is time varying here, as the Lyapunov candidate and find the largest $\rho(t)$ in the interval$[t_0,t_f]$ using SoS programming and binary search for $\rho(t)$ as in \cite{lqrtree}, which gives us the region

\begin{align}
 \mathcal{B}(\rho(\cdot),t) = \{\textbf{x}|0\le V(\textbf{x},t) \le \rho(t)   \}
\end{align} where $V$, the value function of the closed loop system, is nothing but $J^*$, the optimal cost-to-go from Eq. \ref{opt_cost}.
 
This must satisfy Eq. \ref{fun}. Similarly for the goal region, the region of attraction is 
\begin{align}
 \mathcal{B}^l(\rho^l(\cdot),t) = \{ \textbf{x}|0 \le V(\textbf{x},t)\le \rho^l(t) \}
\end{align}
where $\rho^l(t)$ represents a constraint on the final value, $\rho(t_f)$ such that $\rho(t_f) \le \rho^l(t)$.
For a time varying system, it is not reasonable to talk about asymptotic stability as this can only be defined for the system as time goes to infinity. However, we can still say that the cost to go function is going downhill and the system is converging to the trajectory for the duration of the trajectory.
The bounded final value condition can be verified by proving that $\mathcal{B}(\rho(.),t) $ is closed over $t \in [t_0,t_f]$. The set is closed if $\forall t \in [t_0,t_f]$ we have,
\begin{align}
V(\textbf{x},t)&\ge 0\ ,\ \ \ \ \forall \textbf{x} \in  \mathcal{B} (\rho(\cdot),t)\label{con1}\\
\dot{V}(\textbf{x},t) &\le  \dot{\rho}(t)\ , \ \forall \textbf{x}  \in \mathcal{B}^\#(\rho(\cdot),t)
\label{con2}
\end{align}where $\mathcal{B}^\#$ is the boundary of the funnel $\mathcal{B}$,

\begin{align}
\mathcal{B}^\#(\rho(\cdot),t) = \{ \textbf{x}|V(\textbf{x},t) = \rho(t)    \}
\end{align}

The first condition (Eq. \ref{con1}) is satisfied by the LQR derivation which makes sure that $\textbf{S}(t)$ is positive definite. The time derivative of the Lyapunov function is given by,
\begin{align}
\dot{J}^*(\bar{\textbf{x}},t) =  2\bar{\textbf{x}}^T\textbf{S}(t)f(\textbf{x}_0(t)+\bar{\textbf{x}},\textbf{u}_0(t)-\textbf{K}(t)\bar{\textbf{x}})+\bar{\textbf{x}}^T\dot{\textbf{S}}(t)\bar{\textbf{x}}    
\end{align}
Tedrake verifies the second condition (Eq. \ref{con2}) by formulating a series of sums-of-squares feasibility programs just as in original LQR Tree algorithm \cite{lqrtree}.
% \begin{multline}
% \hat{\dot{j}}^*(\bar{\textbf{x}},t) - \dot{\rho}(t)+h_1(\bar{\textbf{x}},t)\big(\rho(t)-\hat{j}^*(\bar{\textbf{x}},t)\big)\\
% +h_2(\bar{\textbf{x}},t)(t-t_k)+h_3(\bar{\textbf{x}},t)(t_{k+1}-t)\le0
% \label{feas}
% \end{multline}
% \begin{align}
% h_1(\bar{\textbf{x}},t)&=\textbf{h}_1^T\textbf{m}(\bar{\textbf{x}},t)\label{h1}\\
% h_2(\bar{\textbf{x}},t)&=\textbf{m}^T(\bar{\textbf{x}},t)\textbf{H}_2\textbf{m}(\bar{\textbf{x}},t),\  \ \textbf{H}_2 = \textbf{H}_2^T>0,\label{h2}\\
% h_3(\bar{\textbf{x}},t)&=\textbf{m}^T(\bar{\textbf{x}},t)\textbf{H}_3\textbf{m}(\bar{\textbf{x}},t), \ \ \textbf{H}_3 = \textbf{H}_3^T>0,\label{h3}
% \end{align}In feasibility programs mentioned in \ref{feas}, the term with $h_1$ is positive from Eq.\ref{h1}. $H_2\   \ and \  \ H_3$ are positive definite from Eq.\ref{h2} and Eq.\ref{h3} which means the terms with $h_2, h_3$ are positive. So now if we find a $H_i$ which satisfies the constraints, the second condition Eq.\ref{con2} will be verified.  

Building a funnel around the required goal trajectory gives the system a little breathing area. It is impossible for a system to track the trajectory obtained from direct collocation, as it will be comprised of cubic splines. It is however reasonable to ask for the dynamics of the system to evolve in such a manner that it lies within the volume defined by the LQR funnels.

% Moreover for a hybrid system, such as the compass gait, the trajectory obtained by direct collocation can have more than one mode. This means that there are discrete jumps in the state space trajectory as the mode changes. In such cases, each piecewise continuous spline is treated as a separate trajectory and then stabilized.

\begin{algorithm}
\caption{CostToGoLimitCycle$(\textbf{x}_0^{l}, \textbf{u}_0^l, \textbf{Q}, \textbf{R})$    \textbf{Sec. IV.A}}\label{lim}
\begin{spacing}{1.3}
\begin{algorithmic}[1]
\State $\textbf{Q}_f$ = \textbf{Q}
\State $converged \gets false$
\While {\textit{not converged}}
\State $[c, \textbf{S}] \gets \texttt{tvlqr}^*(\textbf{x}_0^l, \textbf{u}_0^l, \textbf{Q}, \textbf{R}, \textbf{Q}_f)$ \{Eqn.\ref{riccati}\}
\State $\textbf{Q}_{fv} \gets \textbf{Q}_f$
\State $\textbf{S}(t_f^{+}) \gets \textbf{S}(0)$
\State $\textbf{x}_f \gets \textbf{x}_0^{1}(t_f)$
\State $\textbf{A}_{cd} \gets \frac{\partial{\textbf{CD}^{pq}}}{\partial{x}}\big|_{\textbf{x}_f}$ 
\State $\textbf{S}(t_f^{-}) \gets \textbf{A}_{cd}^{T} \textbf{S}(t_f^+) \textbf{A}_{cd}$ \{Eqn. \ref{jump}\}
\State $\textbf{Q}_f \gets \textbf{S}(t_f^-)$
\If {$\mid \mid \textbf{Q}_f - \textbf{Q}_{fv} \mid \mid_F \ < \ threshold$}
\State $converged \gets true$
\EndIf
\EndWhile
\State \Return [c,\textbf{S}]
\end{algorithmic}
\end{spacing}
\end{algorithm}
\subsection{Estimation of the nominal limit cycle}

%The compass gait system has a stabilizable limit cycle behavior executing the trajectory $\textbf{x}_0^l(t)$ and $\textbf{u}_0^l(t)$. That is, for some bounded perturbation $< \rho^l(t)$, the system stays inside the bounded region around the limit cycle trajectory. Dynamics of compass gait can be factored into modes which are identical and the discrete transition is only due to resetting dynamics to an initial state. That is, a guard event only resets the system in the current mode (in dynamics sense) with another configuration. 

The hybrid system under our consideration has a stabilizable limit cycle behavior executing the trajectory $\{ \textbf{x}_0^l(t)$, $\textbf{u}_0^l(t) \}$. That is, for some bounded perturbation $< \rho^l(t)$, the system stays inside the bounded region around the limit cycle trajectory. Dynamics of a hybrid system can be factored into modes and the discrete transition is only due to resetting the state governed by the dynamics of current mode to an initial state governed by dynamics of the next mode. That is, a guard event only resets the system in the current mode (in dynamics sense) with another configuration.

%For such a system designing a controller has its own region of attraction beyond which we have no guarantee for stabilizing the state back to the region of interest. Hence we first create sample trajectories that drive the system from such states outside the region of interest back to the limit cycle. We construct a tree like RRT where the edges are nominal trajectories obtained using the direct collocation trajectory optimization. We use TVLQR controllers to stabilize the system at these trajectories. The aim is to have a plan for any state that has a feasible trajectory that eventually leads to the limit cycle. 

%The time instances just before and after this transition or collision event is $t^-$ and $t^+$ and the states $\textbf{x}(t^-)$ and $\textbf{x}(t^+)$ respectively. Let the collision dynamics function $\textbf{CD}$ be such that $\textbf{x}(t^+) = \textbf{CD}(\textbf{x}(t^-))$ (Eqn. \ref{cd1} \& \ref{cd2}). For a point to point trajectory optimization with $k$ knot points, $\textbf{x}_1,...,\textbf{x}_k$, we find single continuous trajectory with a constraint $\textbf{x}_1 = \textbf{CD}(\textbf{x}_k)$ to find a periodic limit cycle trajectory for the system. 

The time instances just before and after this transition or collision event are $t^-$ and $t^+$ and the states are $\textbf{x}(t^-)$ and $\textbf{x}(t^+)$ respectively. Let the collision dynamics function $\textbf{CD}^{pq}$ which takes you from mode $p$ to $q$ be such that $\textbf{x}(t^+) = \textbf{CD}^{pq}(\textbf{x}(t^-))$ (Eqn. \ref{cd1} \& \ref{cd2}). To estimate the periodic limit cycle trajectory of the hybrid system we perform point to point trajectory optimization with $k$ knot points, $\textbf{x}_1,...,\textbf{x}_k$, with the constraint that $\textbf{x}_1 = \textbf{CD}^{pq}(\textbf{x}_k)$ which gives a set of continuous trajectories.

A hybrid system might have more than one stable limit cycle. This is attributed to the range of different possible inputs -- all of which achieve limit cycle stability. We choose the limit cycle that results in a local optimum given the cost function $\int_0^{\infty}\textbf{u}^T(t)\textbf{R}\textbf{u}(t) dt$. Although we can design an LQR-tree algorithm which finds a policy from any initial state to the region of attraction of this open loop limit cycle, it is important to design a TVLQR controller which stabilizes the system around this limit cycle itself to increase the region of attraction of the open loop system. This region of attraction is defined by ${\bar{\textbf{x}}}^{T}(t)\textbf{S}^l(t){\bar{\textbf{x}}}(t) \le \rho ^l(t)$ as discussed in section \ref{sec:tvlqr} and IV-B

% Unlike passive dynamic walking compass gait, a dynamic walker with input on flat terrain has more than one stable limit cycle. This is attributed to the range of different possible inputs -- all of which achieve limit cycle stability. We choose the limit cycle that results in a local optimum given the cost function $\int_0^{\infty}\textbf{u}^T(t)\textbf{R}\textbf{u}(t) dt$ (obtained as discussed in the next paragraph). Although we can design an LQR-tree algorithm which finds a policy from any initial state to the region of attraction of this open loop limit cycle, it is important to design a TVLQR (Time-Varying Linear Quadratic Regulator) controller which stabilizes the system around this limit cycle itself to increase the region of attraction of the open loop system. Estimation of this region of attraction is discussed in the coming sections.

 \begin{figure}
    \centering
    \includegraphics[width=0.5\textwidth]{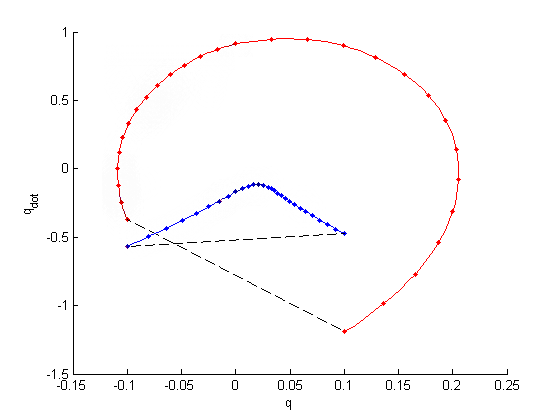}
    \caption{The limit cycle, $x_0^l(t)$, is estimated using direct collocation for hybrid systems. The dotted points show the knot points used by direct collocation between which it fits cubic splines. The red line represents the swing leg and the blue line the stance leg \cite{drake}.}
    \label{phaseplot}
\end{figure}

 \begin{algorithm}
 \caption{LQRTreesAroundLimitCycle(\textbf{Q},\textbf{R})}\label{lqrtrees}
 \begin{spacing}{1.3}

 \begin{algorithmic}[1]
 \State $[\textbf{x}_0^l(t), \textbf{u}_0^l(t)] \gets \texttt{PeriodicDirCollocation}^*()$ %\textbf{Sec. IV.C}
 \State $[c, \textbf{S}] \gets CostToGoLimitCycle(\textbf{x}_0^l, \textbf{u}_0^l)$
 \State $[V,\rho] \gets \texttt{FTV}^*(\textbf{x}_0^l, \textbf{u}_0^l, \textbf{S}, c)$ 
 
 \State $T.init(\{\textbf{x}_0^l, \textbf{u}_0^l,V,\rho,NULL\})$
 \State $converged \gets false$
  \State $iter \gets 0$
 \While {\textit{not converged}}
 \State $\textbf{x}_s \gets $ Uniform Random Sample from State Space  
 \State $\textbf{x}_{near} \gets $ Nearest Neighbor with CostToGo metric
  \State $parent \gets $ Pointer to node containing $\textbf{x}_{near}$
  \If {$\textbf{x}_s$ in $\texttt{RegionOfAttraction}^*$($\textbf{x}_{near}$)}
  \State $iter \gets iter+1$
  \State continue
  \EndIf
 \State $[\textbf{x}_{0},\textbf{u}_{0},Success] \gets \texttt{DirCollocation}^*(\textbf{x}_{near},\textbf{x}_{s})$ %\textbf{Sec. IV.D}
 
 \If {$Success$}
 \State $iter \gets 0$
 \State $V_{near} \gets$ region of attraction of $parent$ 
 \State $[c,\textbf{S}]\gets\texttt{tvlqr}^{*}(\textbf{x}_{0},\textbf{u}_{0},\textbf{Q},\textbf{R},V_{near})$ \textbf{Sec. IV.A}
 
 \State $[V,\rho] \gets \texttt{FTV}^*(\textbf{x}_{0},\textbf{u}_{0},\textbf{S},c,V_{near})$ \textbf{Sec. IV.B}
 
 \State $T.add(\{\textbf{x}_{0},\textbf{u}_{0},V,\rho,n\})$
 \Else
 \State $iter \gets iter+1$
 \EndIf
 
 \If {$iter \ge MAXITER$}
 \State $converged \gets true $         
 \EndIf

% \If {$\mid \mid Q_f - Q_{fv} \mid \mid_2 $}
% \State $converged \gets true$
% \EndIf
 \EndWhile
 \end{algorithmic}
 \end{spacing}
 \end{algorithm}

\subsection{Direct collocation for hybrid systems} \label{sec:dircol}
% in this para should we address the mode sequence prob and posa in future work in final report?
We use direct collocation to give locally optimal trajectories from a given pose to the goal position. This is required as we need a trajectory that connects a newly sampled state to the nearest funnel in the TVLQR cost function sense. For the hybrid trajectory optimization, we add the mode transition constraint in the optimization problem. The transition can be easily described using the collision event which occurs in the hybrid dynamical system. We search for this collision event between knot points so that the system dynamics at this event is accounted for. Once this transition is added into the dynamic model, a normal use of the collocation gives us the required nominal trajectory, $\textbf{x}_0(t)$ and $\textbf{u}_0(t)$. Another problem to be addressed here is the mode sequence that is to be given to the optimizer. We assume that we know the mode sequence of our hybrid system and input it with an initial guess for the time when the transitions occur. However, for systems with several modes, the number of possible mode sequences become enormously high. For such systems, we can let the solver figure out the order of contacts using the implicit trajectory optimization method proposed by Posa \textit{et al.} \cite{posa}. 

We cannot directly use the trajectory generated by the optimizer (line 14 in Alg. \ref{lqrtrees}) method as a sample policy since the probability of a given initial state being one of the states in this nominal trajectory, $\textbf{x}_0(t)$ and $\textbf{u}_0(t)$, is zero (a trajectory in state space has zero volume). The TVLQR controller however stabilizes the system around this trajectory and finds the funnel of attraction which has a non-zero volume in state space (and hence has a non-zero probability of occurrence). For a hybrid trajectory, we have piecewise continuous sub-trajectories punctuated by discrete transitions. We apply TVLQR controller to stabilize the system around every piece of continuous trajectories and find the funnel around each trajectory separately. The jump Riccati equation finds a cost to go $S(t^-)$ such that every state inside $\textbf{x}^T(t^-)\textbf{S}(t^-)\textbf{x}(t^-) < \rho(t^-)$ lands in $\textbf{x}^T(t^+)\textbf{S}(t^+)\textbf{x}(t^+) < \rho(t^+)$. This ensures that the system is stable throughout the hybrid trajectory.

 \begin{figure}
    \centering
    \includegraphics[width=0.5\textwidth]{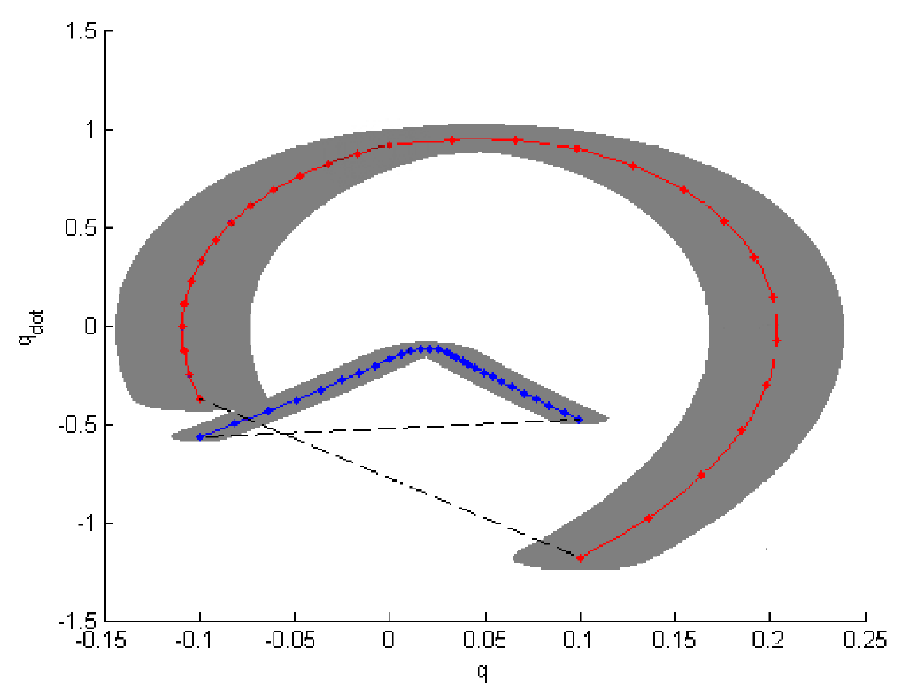}
    \caption{The limit cycle of compass gait. The red and the blue curves are the nominal trajectories of the swing and the stance leg respectively. The grey region shows the verified funnel the nominal limit cycle trajectory trajectory $x_0^l(t)$ under the action of TVLQR controller.}
    \label{roa}
\end{figure}

\subsection{Growing the LQR Trees}
The LQR tree is initialized with the goal funnel of the nominal limit cycle $\textbf{x}_0^l(t)$ and $\textbf{u}_0^l(t)$. Next, a state $\textbf{x}_s$ is randomly sampled uniformly from the state space.  If the newly sampled point falls inside the funnel then it is discarded (Alg.\ref{lqrtrees} line 11). Otherwise, we find the nearest point in the existing tree by using the LQR cost-to-go distance metric\cite{lqrtree}. This in a way means that the controller requires minimum effort to bring the sampled point to this nearest point. Once the closest node in the tree is identified we perform trajectory optimization like direct collocation connecting the sampled point and the nearest point (Alg.\ref{lqrtrees} line 14). Then we design the TVLQR controller for the trajectory. Following this, the funnel around this trajectory under the stabilization of TVLQR is verified using section IV-B. We add the newly verified funnel to the existing tree. Our algorithm terminates when a predetermined number of consecutive sample points returns failure or fall in existing funnels.

One can find the TVLQR controller for the given trajectory using Eq. \ref{optimal_u}. We feed the piecewise continuous trajectory of the hybrid trajectory and design separate TVLQR controllers for each mode. Then we verify every controller separately and determine the Lyapunov function $V(t)$ and the size of the funnel $\rho$. Hence the region of attraction is given by $V(t) \le \rho(t)$ with which we check if a sampled point falls inside a funnel or not.  

\section{EXPERIMENTAL SETUP}

% TO BE ADDED HERE
% Unlike passive dynamic walking compass gait, a dynamic walker with input on flat terrain has more than one stable limit cycle.

\begin{figure}[H]
    \centering
    \includegraphics[width=0.35\textwidth]{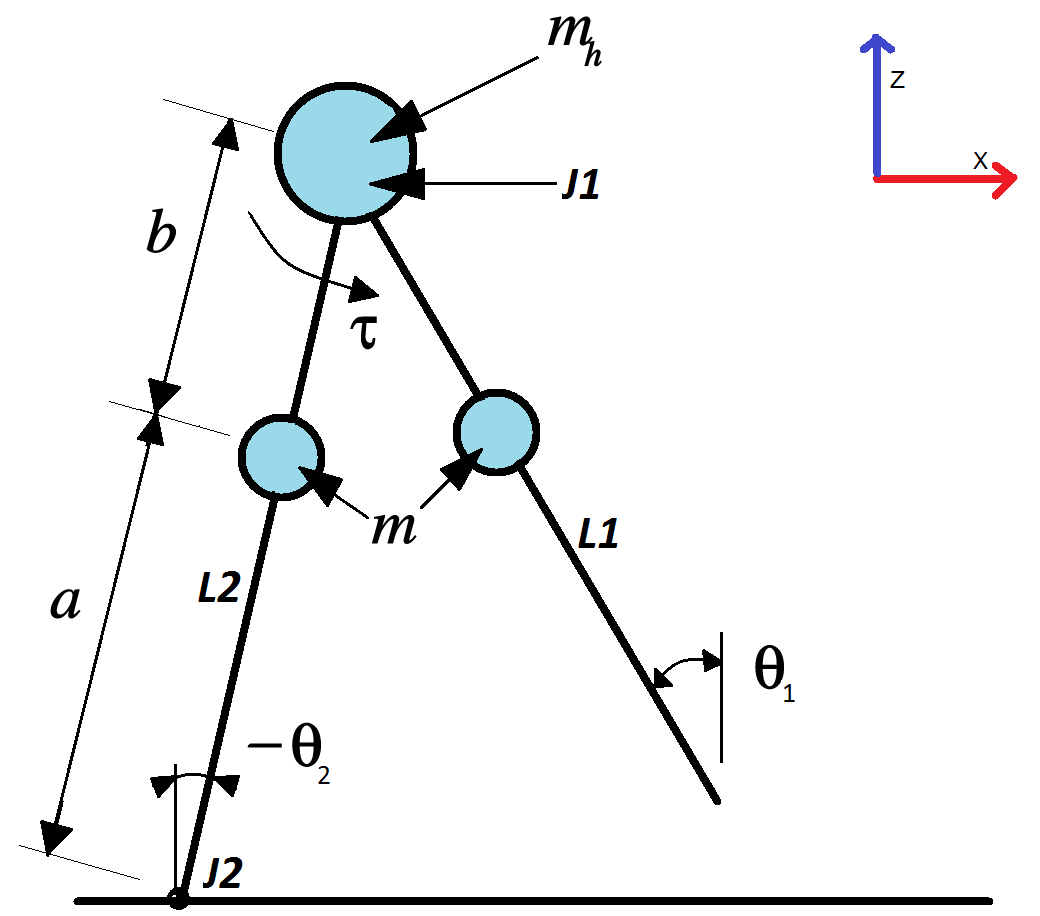}
    \caption{The compass gait. Link $L_2$ is attached to the ground and $L_1$ is attached to $L_2$ by pin joints $J_2$ and $J_1$ respectively. The model has a control input $u$ at joint $J_1$ that exerts torque between the two legs and a mass $m_h$ is attached at this joint.}
    \label{gaitfig}
\end{figure}

The dynamical system in which we investigate our algorithm is a minimalistic version of a compass gait. The compass gait is modeled as a two link ($L_1$ and $L_2$) manipulator with masses $m$ concentrated at their center Fig.\ref{gaitfig}. 

 We define the state of the system \textit{i.e.,} $\theta_1$ for swing ($L1$) and $\theta_2$ for stance ($L2$) and their derivatives with respect to the world frame attached to the ground as shown in Fig.\ref{gaitfig}. The state of the system is given by, $\textbf{x} = [\theta_1, \theta_2, \dot{\theta}_1, \dot{\theta}_2]^T$. The dynamics governing the compass gait model can be described as $\textbf{M}(\textbf{x})\ddot{\textbf{x}} + \textbf{C}(\textbf{x},\dot{\textbf{x}})\dot{\textbf{x}} + \textbf{G}(\textbf{x}) = \textbf{B}u$, where the matrices $M(\textbf{x})$, $\textbf{C}(\textbf{x},\dot{\textbf{x}})$ and $\textbf{G}(\textbf{x})$ contain the inertial, Coriolis and the gravity terms as given in \cite{underactuated}.

\begin{align}
    \textbf{M}(\textbf{x})\ddot{\textbf{x}} + \textbf{C}(\textbf{x},\dot{\textbf{x}})\dot{\textbf{x}} + \textbf{G}(\textbf{x}) = \textbf{B}u
\end{align}
where,

\begin{align}
\textbf{M}=\begin{bmatrix}
mb^2 & -mlb \ cos(\theta_2 - \theta_1)\\ 
-mlb \ cos(\theta_2 - \theta_1) &  (m_h+m)l^2 + ma^2
\end{bmatrix}
\end{align}

\begin{align}
\textbf{C} &=\begin{bmatrix}
0 & mlb \ sin(\theta_2 - \theta_1)\ \dot{\theta_2}\\ 
mlb \ sin(\theta_2 - \theta_1)\ \dot{\theta_1} &  0
\end{bmatrix}
\end{align}

\begin{align}
\textbf{G} &=\begin{bmatrix}
 mgb \ sin(\theta_1)\\ 
-(m_hl + ma + ml)g \ sin(\theta_2)
\end{bmatrix}
\end{align}
where $l=a+b$. The compass gait is placed on a flat terrain, \textit{i.e.,} the acceleration due to gravity $g$ is in the $-\hat{z}$ direction. During the walking cycle, when the swing leg collides with the ground we generally assume conservation of angular momentum \footnote{We do not consider the change in the mechanical energy of the system during this conservation. In \cite{hal}, they prove that $\Delta E$ is always negative.} about the hip joint and toe of the swing leg. The pre-impact and post-impact state parameters can be linearly related using the collision dynamics as shown below \cite{underactuated},\begin{align}
\begin{bmatrix}
\dot{\theta_1}^+ \\
\dot{\theta_2}^+
\end{bmatrix} = (\textbf{T}^+)^{-1}\textbf{T}^- \begin{bmatrix}
\dot{\theta_1}^- \\
\dot{\theta_2}^-
\end{bmatrix}
\label{cdtrans}
\end{align}

\begin{align}
\textbf{T}^+ = \begin{bmatrix}
mb(b-l\ \cos \gamma) &  ml(l-b\ \cos \gamma) + ma^2 + m_hl^2\\ 
mb^2 & -mbl\ \cos \gamma
\end{bmatrix}
\label{cd1}
\end{align}

\begin{align}
\textbf{T}^- = \begin{bmatrix}
 -mab & -mab + (m_hl^2 + 2mla)\ cos(\gamma)\\ 
0 & -mab
\end{bmatrix}
\label{cd2}
\end{align}where $\dot{\theta}_1^-,\dot{\theta}_2^-$ and $\dot{\theta}_1^+,\dot{\theta}_2^+$ are the joint angular velocities just before and after the collision, $\gamma = \theta_1 - \theta_2$,  $\textbf{T}^-$ and $\textbf{T}^+$ are the transition matrices that contain the coefficients of conservation of angular momentum. The size of $\textbf{T}^-$ and $\textbf{T}^+$ matrices are $2\times2$ because angular momentum is conserved about two points and for two angular velocities. During collision event, the mapping between joint angles before and after collision is obtained by merely interchanging them \textit{ i.e.,} $\theta_1^+ = \theta_2^-$ and $\theta_2^+ = \theta_1^-$. The complete collision dynamics $\textbf{CD}(\textbf{x})$ is found using this and velocity mappings in Eq. \ref{cdtrans}. 

%In case of compass gait, the collision dynamics function becomes $\textbf{CD}(\textbf{x}) = (\textbf{T}^+)^{-1}\textbf{T}^-$ (Eq. \ref{cdtrans}).

%The compass gait exhibits a dynamic limit cycle walking behavior on a flat terrain. The aim is to have a plan for any state that has a feasible trajectory which eventually leads to the limit cycle.

% For such a trajectory designing a controller has it's own region of attraction beyond which we have no guarantee for stabilizing a state back to the region of interest. Hence we first create sample trajectories that drives the system from such states outside the region of interest back to the limit cycle. We construct a tree like RRT where the edges are nominal trajectories obtained using the direct collocation trajectory optimization. We use TVLQR controllers to stabilize the system at these trajectories

\section{EXPERIMENTS AND RESULTS}

\subsection{Compass Gait Model}
The compass gait model whose dynamics are described above is simulated with the following parameters. The mass at the hip $m_h=10kg$, link masses $m=5kg$, link lengths $a=b=0.5m$ and $g=9.8m/s^2$. Here $\textbf{x}= [\theta_1, \theta_2, \dot{\theta_1}, \dot{\theta_2}]$ and $\textbf{u}= [\tau]$ at the hip. The joints are assumed to be frictionless and the collisions are inelastic. 

\subsection{Direct Collocation and LQR Trees}
The direct collocation procedure is a nonlinear trajectory optimization problem which requires an initial guess for $\textbf{x}(t)$ and $\textbf{u}(t)$. The initial guess for $\textbf{x}(t)$ is a straight line connecting the random sample and the nearest point in the existing tree based on the TVLQR cost function. The initial guess for input trajectory $\textbf{u}(t)$ is a random value. The mode transition constraint is given by the collision dynamics (Eq. \ref{cdtrans}) which occurs in the compass gait model when the distance between the swing leg and the ground is zero. The direct collocation function has very sparse gradients and the constraints depend upon the values at knot points or adjacent knot points. Solvers such as SNOPT \cite{snopt} can very efficiently solve such nonlinear programs with sparse gradients. Every iteration of direct collocation is terminated and a new state is sampled if there is no result after 40 seconds or if the solver fails. 

\begin{figure}
     \begin{center}
        \subfigure[]{%
            \label{query1}
            \includegraphics[width=0.5\textwidth]{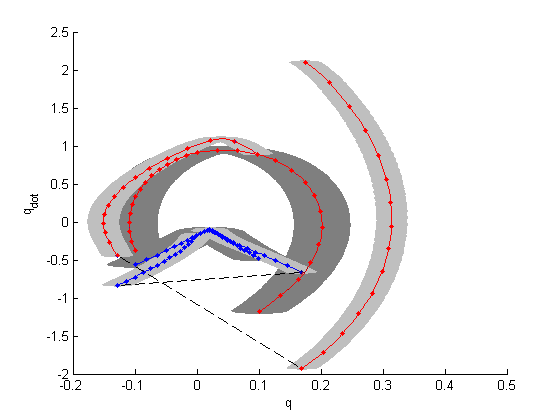}
        } \\
        \subfigure[]{%
           \label{query2}
           \includegraphics[width=0.5\textwidth]{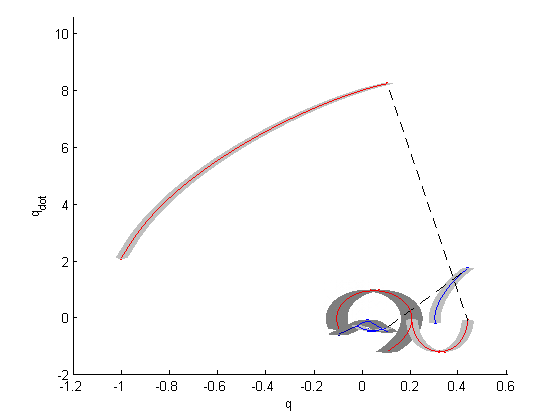}
        }%  ------- End of the first row ----------------------%
%         \subfigure[Caption of Third Figure]{%
%             \label{fig:third}
%             \includegraphics[width=0.4\textwidth]{ThirdFigure}
%         }%
%         \subfigure[Caption of Fourth Figure]{%
%             \label{fig:fourth}
%             \includegraphics[width=0.4\textwidth]{FourthFigure}
%         }%
% %
    \end{center}
    \caption{%
        Phase plane plot showing the recovery of the compass gait with (a) velocity perturbation only (b) both velocity and position perturbation. The curves in red and blue show the nominal trajectory of the swing and the stance legs respectively. The gray region shows the verified funnel around each trajectory under the action of TVLQR controller. 
     }%
   \label{fig:subfigures}
\end{figure}

The parameters of the LQR tree algorithm are $\textbf{Q} = diag([10,10,1,1])$, $\textbf{R} = [15]$. We terminate after 500 consecutive samples (MAXITER om Alg. \ref{lqrtrees}) fall in the existing tree or fail to find a trajectory to the tree. The LQR tree is considered to have reasonably covered the entire region of stabilizable state space for compass gait if 500 consecutive samples either fall inside the funnels or outside and the direct collocation fails.  
\begin{center}
%\begin{tabular} { | m{3.5cm} | m{1.5cm} | m{1.2cm} | m{1.3cm}|}
\begin{tabulary}{9cm}{|C|C|C|C|}
 \hline
 System & Pendulum & Cartpole & Compass gait\\
 \hline
 No. of Samples & \textcolor{blue}{146} & \textcolor{blue}{-} & \textcolor{blue}{252}\\
& \textcolor{red}{6068} & \textcolor{red}{5821} &\textcolor{red}{-} \\
 \hline
 No. of nodes & \textcolor{blue}{146} & \textcolor{blue}{-} & \textcolor{blue}{192}\\
 & \textcolor{red}{477} & \textcolor{red}{881} & \textcolor{red}{-}\\
 \hline
 Time taken for  & \textcolor{blue}{0h 2m} & - & \textcolor{blue}{6h 21m}\\
planning & \textcolor{red}{0.5h} & \textcolor{red}{2h} & \textcolor{red}{-} \\
 \hline
\end{tabulary}
\label{results}
\captionof{table}{\\Results for pendulum and cart pole systems obtained from \cite{lqrtree} and \cite{lqrtree2}. The results of verification by SoS optimization and simulation are shown in blue and red fonts respectively. \\ \textbf{Note:} A ``-" denotes unavailability of data from \cite{lqrtree} or \cite{lqrtree2}.}\end{center}

The goal limit cycle trajectory $\textbf{x}_0^l(t)$ and $\textbf{u}_0^l(t)$ are obtained directly from the \texttt{PeriodicDirCollocation}() method. Also, a TVLQR controller given a nominal trajectory can be found using \texttt{tvlqr}(). The \texttt{sampledFiniteTimeVerification}() (aka \texttt{FTV}) finds the ratio of value of Lyapunov function $V(\textbf{x})$ to the size of the funnel $\rho(t)$. So to check if a point $\textbf{x}$ is in a given funnel, we just need to check $V(\textbf{x})/\rho(t) \le 1$.

All the function implementations denoted by the \texttt{typewriter} typeface and asterisk (*) superscripts in the Alg. \ref{lim} \& \ref{lqrtrees} are available in \texttt{DRAKE} \cite{drake}.

% the \texttt{FTV} which is the \texttt{sampledFiniteTimeVerification}() available in \texttt{DRAKE}. 
%tvlqr function in drake finds the controller around a given trajectory
%sampledTimeVerification returns the V(t)/ p(t)

\subsection{Experiments with Initial Perturbation and Noise}

Once the state space is filled with LQR trees the compass gait has a recovery policy to stabilize itself back to the limit cycle's region of attraction from any arbitrary and stabilizable initial condition. It is to be noted that not all states of the compass gait are stabilizable no matter what input we apply. For example, a compass gait lying down cannot pump energy to get itself up in any way to get back up on two legs. We tested our algorithm by perturbing both the position and velocity states of the system. The figure \ref{query1} shows the phase portrait of the case in which there was an initial velocity perturbation and figure \ref{query2} shows the phase portrait of the case in which there was both velocity and position perturbation. Both the above cases are simulated by considering a white Gaussian noise in the forward simulation with 0.05  as standard deviation.

\subsection{Inferences}
TABLE I provides a comparison of the two methods to generate LQR trees and the dynamical systems they were applied to. It took around 381 minutes to completely cover the state space\footnote{It was evaluated on a Intel Core i7-6700K 4 GHz Quad-Core Processor with 32 GB RAM} with funnels for the compass gait system being evaluated. The results reflect our intuition in that the simulation methods take longer to build the LQR trees. For the compass gait, 23\% of the time was used up for direct collocation and 58\% for funnel verification. Trajectory verification takes a big chunk of the total time because the trajectories being verified are hybrid trajectories. There are discrete jumps in the trajectories as the mode of the system changes with time. The start of the next mode trajectory is obtained using the jump Riccati equation. 

%TBR: redundant last line above

% \begin{figure}
%     \centering
%     \includegraphics[width=.44\textwidth]{query1c}
%     \caption{Phase plane plot for the first case with velocity perturbation }
%     \label{query1}
% \end{figure}
% \begin{figure}
%     \centering
%     \includegraphics[width=.44\textwidth]{query2}
%     \caption{Phase plane plot for the second case with velocity and position perturbations }
%     \label{query2}
% \end{figure}

% \begin{figure}[ht!]
% \centering
%   \begin{subfigure}{\textwidth}
%   \includegraphics[width=.5\textwidth]{query1c}
%     \caption{}
%   \label{query1} 
% \end{subfigure}

% \begin{subfigure}{\textwidth}
%   \includegraphics[width=.5\textwidth]{query2}
%     \caption{}
%   \label{query2}
% \end{subfigure}
% \caption[Two numerical solutions]{(a) Phase plane plot for the first case with velocity perturbation  (b) Phase plane plot for the second case with velocity and position perturbations }
% \end{figure}

\section{FUTURE WORK}

% From the results, we can see that the LQR-tree \cite{lqrtree} algorithm scales poorly for hybrid systems and also with the increase in a number of degrees of freedom. 

One of the interesting future works would be to extend the algorithm for the case of a high DoF compass gait (Compass Gait with knees\cite{kneed} -- 3 DoFs). We have a more efficient walking limit cycle with knees than keeping the knees straight.  To find a policy for kneed gait with LQR trees of kneeless gait, a set of approximations could be made to map a given state in kneed gait to a state in simple gait and follow its policy given by the LQR-tree. Also the number of possible mode sequences becomes intractable for high DoF systems. For such systems (kneed compass gait) mode sequence specification could be circumvented by using the algorithm laid by \cite{posa}.

% Since the compass gait with knees has numerous combinations of mode sequence it is not possible to specify a sequence a priori as we had done for the low DoF system. To overcome this one can use the algorithm laid out by \cite{posa} which just adds the contact forces as decision variables of the trajectory optimization problem in addition to state and input vector at the knot points and subject it to additional complementarity constraints. 

%The various approximations that we would like to examine are,

% \begin{itemize}

% \textbf{Straight and Locked Knee:}
When a perturbation causes the kneed compass gait to go off the region of attraction of its limit cycle we can approximate the instantaneous configuration by 1) making the links move to a configuration with equal angles at the knees and by locking the knee angles 2) stretching the knee to obtain an equivalent low DoF configuration. In both cases, we go to the described configurations using direct collocation with the constraint that the knee angles are equal (or zero for straight knees). Once we stabilize the system with the knees straight or locked, the system will exhibit a stable kneeless limit cycle. We just need one trajectory from one of the points of this limit cycle to a point in the more efficient kneed limit cycle. Moreover, we can use these trajectories obtained from such mappings as seed trajectories to build trees in the state space of kneed compass gait.

% \item \textbf{PCA for high DoF configuration}: Find the redundant component of the state in for the high DoF compass gait and plan in the reduced dimension space. This analysis is specific to the trajectory from which we sample the states.

% \end{itemize}

\section{CONCLUSIONS}

We presented a method to use LQR trees to control hybrid systems like compass gait. We also devised a method (Algorithm.\ref{lim}) to find the region of attraction of TVLQR controller around a dynamic limit cycle which is important to find the funnel around every other trajectory of LQR Tree. This method effectively covers the stabilizable regions in a bounded state space with a series of feedback stabilized sample trajectories that lead to the goal trajectory. The algorithm was tested for a compass gait modeled as a two-link manipulator. We also discuss how an LQR Tree built for a lower DoF robot may be reused on a higher DoF systems. While we probabilistically cover stabilizable regions of state space to provide a policy beyond a controller's region of attraction, it takes more than six hours to do this for one goal trajectory. For a new goal trajectory, one can reuse the tree if a trajectory can be found between any two points -- one in the old and one in the new goal trajectory. Otherwise, it could potentially take another six hours of planning. 

% Otherwise another 6 hours of planning is needed. 

% \begin{figure}
%     \centering
%     \includegraphics{knee}
%     \caption{Simulation of a kneed compass gait in DRAKE. The model has point masses attached to the center of each link.}
   
%     \label{knee}
% \end{figure}

\addtolength{\textheight}{-12cm}   % This command serves to balance the column lengths
                                  % on the last page of the document manually. It shortens
                                  % the textheight of the last page by a suitable amount.
                                  % This command does not take effect until the next page
                                  % so it should come on the page before the last. Make
                                  % sure that you do not shorten the textheight too much.

%%%%%%%%%%%%%%%%%%%%%%%%%%%%%%%%%%%%%%%%%%%%%%%%%%%%%%%%%%%%%%%%%%%%%%%%%%%%%%%%

%%%%%%%%%%%%%%%%%%%%%%%%%%%%%%%%%%%%%%%%%%%%%%%%%%%%%%%%%%%%%%%%%%%%%%%%%%%%%%%%

%%%%%%%%%%%%%%%%%%%%%%%%%%%%%%%%%%%%%%%%%%%%%%%%%%%%%%%%%%%%%%%%%%%%%%%%%%%%%%%%

\section*{ACKNOWLEDGMENT}

 We specially thank Prof. Russ Tedrake for his quick responses to our queries regarding \texttt{DRAKE} simulator \cite{drake}. We also thank Prof. Dmitry Berenson for his motivation behind exploring this area of research.

\end{document}